\documentclass{mva_style}
\usepackage{graphicx}
\usepackage{tikz}
\usepackage{amssymb,amstext,amsmath}
\usepackage{sidecap}
\usepackage{url}
\usepackage[sort]{cite}

\usepackage{hyperref}
\hypersetup{
    colorlinks,%
    citecolor=blue,%
    filecolor=black,%
    linkcolor=red,%
    urlcolor=black
}

\makeatletter
\def\hlinewd#1{%
\noalign{\ifnum0=`}\fi\hrule \@height #1 %
\futurelet\reserved@a\@xhline}
\usepackage{tikz}
\usetikzlibrary{shapes,arrows}
\usetikzlibrary{trees}
\usepgflibrary{decorations.fractals}
\usepgflibrary{decorations.footprints}
\usetikzlibrary{petri}
\usepackage{colortbl}
\usepackage{xcolor}
\usepackage[all]{xy}
\usetikzlibrary{arrows,decorations.pathmorphing,backgrounds,fit,positioning,shapes.symbols,chains}

\finalcopy 

\begin{document}
\title{Client-Driven Content Extraction Associated with Table}

\author{
   K.C. Santosh and Abdel Bela\"id\\
  LORIA - Universit\'e de Lorraine\\
  54506 Vandoeuvre-l\`es- Nancy, France\\
  {\tt \{santosh.kc, abdel.belaid\}@loria.fr}\\
}

\maketitle

\section*{\centering Abstract}
\textit{
The goal of the project is to extract content within table in document images based on learnt patterns. Real-world users i.e., clients first provide a set of key fields within the table which they think are important. These are first used to represent the graph where nodes are labelled with semantics including other features and edges are attributed with relations. Attributed relational graph (ARG) is then employed to mine similar graphs from a document image. Each mined graph will represent an item within the table, and hence a set of such graphs will compose a table. We have validated the concept by using a real-world industrial problem.  
}

\section{Introduction}\label{sec: intro}
In document analysis and processing, table extraction from document images has been received an important attention since it contains key information. In the context of table extraction~\cite{zanibbi04IJDAR,Couasnon:06:IJDAR,Hurst:06:IJDAR,EmbleyHLN:06:IJDAR}, document image analysis and processing basically describes table either in terms of lines and (un)analysed text blocks, a set of cells resembling the two-dimensional grid or a set of strings that are integrated with each other via relations, for instance. 

Basically, table detection and its structure recognition are two major tasks. Table detection can be taken as a primary issue, which is however does not provide a complete solution~\cite{MandalCDC:06:IJDAR} since one needs to be able to extract key fields within it. Existing methods such as table segmentation~\cite{LiangWS11} do not extract key fields, nor do they explicitly perform the content understanding~\cite{DeckertSEG11}. Note that structural information by considering relations between the contents, for instance can be very useful in indexing and retrieving document information~\cite{Couasnon:06:IJDAR}. To analyse table-forms structure, rulings techniques are basically limited without a priori knowledge about table organisation~\cite{zanibbi04IJDAR}. Such concepts are completely failed since not all tables possess graphical lines. Besides, plain ascii texts, text blocks are used. Detecting columns, lines and headers, and representing them in terms of graph, for instance is interesting since it contains structural information. In order to fully exploit table in the scanned documents rather than just outlining the overall boundary, it is  interesting to extract those fields that are important or meaningful for the clients. To handle this, in this paper, key fields are provided by the clients. These key fields are then used to build a graph so that it can be applied for table extraction in the absence of clients. 

The rest of the paper is organised as follows. We start with explaining the proposed method in Section~\ref{sec:prop}. Full experiments are reported and analysed in Section~\ref{sec:expe}. The paper is concluded in Section~\ref{sec:conc}.

\section{Proposed method}\label{sec:prop}
Generally speaking, table is composed of similar items (sometimes just a single) even when columns alignment and corresponding text flow (either in a single or multiple lines) are not guaranteed. Given an input pattern (i.e., an item, for instance) from a client, finding similar patterns from the document is the core part of the paper. It not only extracts important fields (in accordance with the client) but also configures table represented by a set of similar patterns. To handle this, we first represent an input pattern via an ARG and perform graph mining so that similar graphs can be extracted that are structurally and semantically similar. Fig.~\ref{block} shows a screen-shot of the overall idea.

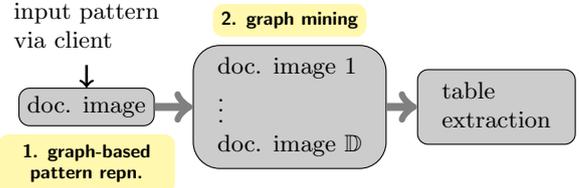
\begin{figure}[tbp]
\centering
\small
\begin{tikzpicture}
[scale = 0.5, auto,
every node/.style={node distance=0.2cm},
comment/.style={rectangle, inner sep= 5pt, text width=4cm, node distance=0.25cm}, 
input0/.style={rectangle, rounded corners, auto}, 
input/.style={rectangle, rounded corners, fill=yellow!40, auto, font=\bfseries\scriptsize\sffamily},
input1/.style={draw, rectangle, rounded corners, fill=gray!40, auto}, 
document/.style={draw, rectangle, rounded corners = 10pt, minimum height=4em, fill=gray!40},
database/.style={ draw, rectangle, rounded corners = -10pt, color = black, minimum height=2em, fill=gray!40}, 
output/.style={rectangle, rounded corners = 1pt, color = black, auto, fill=magenta!40}, 
force/.style={rectangle, rounded corners, draw, fill=black!10}
]
\node [document] (doc) {
\begin{tabular}{l} 
doc. image 1 \\
$\vdots$ \\
doc. image ${\mathbb D}$ \\
\end{tabular} 
};
\node [input1, right =0.4cm of doc] (tableO) {\begin{tabular}{l} 
table\\
extraction
\end{tabular} };
\node [input, above =0.1cm of doc] (graphM) {2. graph mining};
\node [input1, left=0.5cm of doc] (graphP) {doc. image};
\node [input, below=0.1cm of graphP] (graphI) {\begin{tabular}{c} 1. graph-based \\ pattern repn. \end{tabular}};
\node [input0, left=-0.6cm of graphP] (inputP0) {};
\node [input0, above=0.3cm of graphP] (inputP) {\begin{tabular}{l} input pattern \\ via client\end{tabular}};



\path[->,very thick, gray, line width = 3pt] 
(graphP) edge (doc)
(doc) edge (tableO);

\path[->, line width = 1pt]
(inputP) edge (graphP);
\end{tikzpicture} 
\mbox{}\\
\caption{Work-flow showing two consecutive phases: graph-based pattern representation and graph mining, to handle table extraction.}\label{block}
\end{figure}
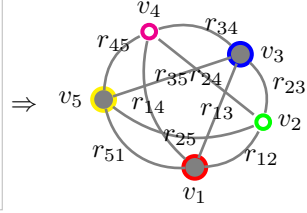
\begin{SCfigure*}[2][t]
\centering
\normalsize
\renewcommand{\tabcolsep}{0.1em}
\begin{tabular}{cc}
\begin{tabular}{c}
\begin{tikzpicture}
\node[-, draw, gray!50] (b) at (0,0) 
{{\includegraphics[scale = 0.45]{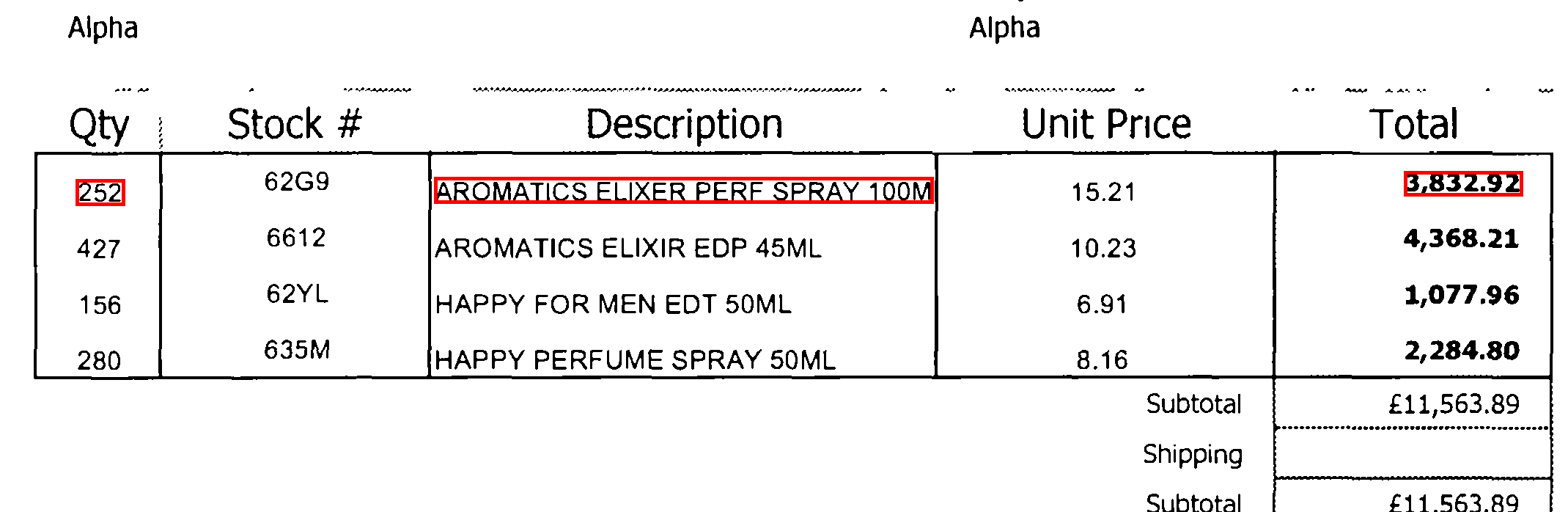}}};
\node at (-2.0, 1.2) {\color{red}{input pattern \textbf{$_\searrow$} }};
%
\end{tikzpicture}
\end{tabular}
 & $\Rightarrow$
 \begin{tabular}{c}
 {
\begin{tikzpicture}[scale=3.0]
\clip (-0.1,0) rectangle (1,1);
\node[] (a) at (0.5,0.2) {};
\node[] (b) at (0.8,0.4) {};
\node[] (c) at (0.7,0.7) {};
\node[] (d) at (0.3,0.8) {};
\node[] (e) at (0.1,0.5) {};
\draw[red, very thick, line width = 1.5pt, fill = gray] (0.5,0.2) circle(0.05cm);
\draw[green, very thick, line width = 1.5pt, fill = white] (0.8,0.4) circle(0.03cm);
\draw[blue, very thick, line width = 1.5pt, fill = gray] (0.7,0.7) circle(0.05cm) ;
\draw[magenta, very thick, line width = 1.5pt, fill =white] (0.3,0.8) circle(0.03cm);
\draw[yellow, very thick, line width = 1.5pt, fill = gray] (0.1,0.5) circle(0.05cm);
\path[-,very thick, gray, line width = 1pt] 
(a) edge [bend right] node[black,right =-1pt]{$r_{12}$} (b)
(b) edge [bend right] node[black, right =-1pt]{$r_{23}$}(c)
(c) edge [bend right] node[black, right = -1pt]{$r_{34}$}(d) 
(d) edge [bend right] node[black, above = -1pt]{$r_{45}$}(e) 
(e) edge [bend right = 40pt] node[black, left = -1pt]{$r_{51}$}(a) 
(a) edge  node[black]{$r_{13}$} (c)
(a) edge   [bend left]node[black]{$r_{14}$} (d)
(b) edge node[black]{$r_{24}$} (d)
(b) edge  [bend left]node[black]{$r_{25}$} (e)
(c) edge  node[black]{$r_{35}$} (e);

\node[below = 4pt] (a) at (0.5,0.2) {$v_1$};
\node[right = 2pt] (b) at (0.8,0.4) {$v_2$};
\node[right =4pt] (c) at (0.7,0.7) {$v_3$};
\node[above =2pt] (d) at (0.3,0.8) {$v_4$};
\node[left =4pt] (e) at (0.1,0.5) {$v_5$};
\end{tikzpicture}
}
\end{tabular}
\end{tabular}
\caption{An example of the input pattern and the corresponding graph that includes missing fields.}\label{missAttribute}
\end{SCfigure*}

\subsection{Graph-based pattern representation}

In any document $d$, the clients provide input pattern(s) while showing the interest of the particular type $t$ of table in either header, body or footer zone: 
$
\mbox{table}_t = \{\mbox{pattern}_n, n\in[1,\mathbb N] \},
$ 
where $\mathbb N$ can be arbitrary. An example of input pattern is shown in Fig.~\ref{missAttribute} i.e., it is just a collection of the selected key fields: $\{ \mbox{field}_i\}_{i =1}^{\mathbb A}$. To represent each field, we define a feature set ${\cal F}$ as $\big\{ \mbox{feature}_f\big\}_{f=1}^{\mathbb F}$. For any $i$-th field, we can formally represent feature as $\mbox{field}_{i}^{\cal F} = \big\{$ 
{\footnotesize
\begin{eqnarray}\label{eq_field}
\renewcommand{\tabcolsep}{0.2em}
\begin{tabular}{ll}
(\texttt{box}: [left, top, right, bottom]); & (\texttt{wSep}: words separation); \\ 
(\texttt{value}: content); & (\texttt{noW}: number of words); \\ 
(\texttt{type}: content type); & (\texttt {noL}: number of lines);\\ 
(\texttt{size}: string length); & (\texttt{label}: \textit{date} and \textit{price}, \\ 
											  & for instance.)$\big\}$ 
\end{tabular}
\end{eqnarray}
}The labels are the derivative of features, representing semantic values via regular expressions. Thanks to the regular expressions, we are able to express a wide range of string values even when we have possible OCR errors due to broken characters and characters are connected with graphics, for instance. To exploit relative positioning between the key fields, we basically use bounding box and its projection into $3 \times 3$ partitions~\cite{Papadias:97:IJGIS} (defined in $IR^2$ i.e., \textit{left}, \textit{right}, $\dots$). For more precision, we integrate the level of neighbourhood $k$ into the basic predefined set of spatial predicates, we have 
{
\begin{eqnarray}
r_{ij} = \textit{spatial\mbox{ }predicate}_{k_1, k_2}(\mbox{field}_i,\mbox{field}_j).
\end{eqnarray} 
}Formally, $k= 0$ for an adjacent (an immediate field), and $k$ varies from $1$ to $\mathbb A-1$ for non-adjacent ones. Note that $k_1$ and $k_2$ represent horizontal and vertical orientations, respectively. 

Now, we introduce a 4-tuple ARG $$G(V, E, F_V, F_E),$$ where
\begin{itemize}\itemsep-0.5em
\item $V$ is a finite set of nodes (fields);
\item $E \subseteq V \times V$ i.e., a finite set of edges and each $r_{ij}\in E$ is a pair of $(v_i,v_j)$ where $v_i,v_j \in V$;
\item $F_{V}: V\rightarrow L_{V}$, $L_V$ represents a set of nodes as well as their labels $\cal{L}$; and 
\item $F_{E}: E \rightarrow R_{E}$, $R_{E}$ represents the edges via relations. 
\end{itemize}
To make graph complete, we also include non-selected fields which are mainly missing and neighbouring fields. To know how many words can be taken for a single field, we simply use intra-field (i.e., maximum distance between the words in a single field) knowledge from the selected key fields. 

\begin{SCfigure*}[5][t]
\centering
\renewcommand{\tabcolsep}{0em}
\begin{tabular}{ccl}
\begin{tabular}{c}
{
\begin{tikzpicture}[scale=3]
\clip (0,0) rectangle (0.801,0.8);
\node[] (a) at (0.2,0.2) {};
\node[] (b) at (0.6,0.2) {};
\node[] (c) at (0.6,0.6) {};
\node[] (d) at (0.4,0.5) {};
\draw[red, very thick] (0.2,0.2) circle(0.03cm);
\draw[green, very thick] (0.6,0.2) circle(0.03cm);
\draw[blue, very thick] (0.6,0.6) circle(0.03cm) ;
\draw[black, very thick] (0.4,0.5) circle(0.03cm);
\path[-,very thick, gray, line width = 1pt] 
(a) edge node[black, below =-1pt]{$r_{12}^q$} (b)
(b) edge node[black, right =-1pt]{$r_{24}^q$}(c)
(b) edge node[black]{$r_{23}^q$}(d) 
(c) edge  node[black, above = -3pt]{$r_{34}^q$} (d) 
(a) edge node[black, above = -7pt]{$r_{13}^q$}(d) 
(a) edge [bend left =70pt] node[black, above = -2pt]{$r_{14}^q$}(c); 
\node[below = 1pt] (a) at (0.2,0.2) { $v_1^q$};
\node[below = 1pt] (b) at (0.6,0.2) {$v_2^q$};
\node[right = 1pt] (c) at (0.6,0.6) {$v_4^q$};
\node[left = 1pt] (d) at (0.4,0.5) {$v_3^q$};
\end{tikzpicture}
} \\ (a) pattern graph $Q$
\end{tabular}
&
$\Rightarrow$

\begin{tabular}{c}
{\footnotesize
{
$\begin{array}{c|cccc}
        & v_1^q        & v_2^q       & v_3^q       & v_4^q\\ \hline 
v_1^q & 0            & r_{12}^q   & r_{13}^q   & r_{14}^q\\[1pt]
v_2^q & r_{21}^q   & 0            & r_{23}^q  & r_{24}^q \\[1pt]
v_3^q & r_{31}^q   & r_{32}^q   & 0           & r_{34}^q \\[1pt]
v_4^q & r_{41}^q   & r_{42}^q   & r_{43}^q  & 0 \\[3pt]
\end{array} $
}
} \\ (b) adjacency matrix
\end{tabular}

&
$\Rightarrow$
\begin{tabular}{c}
{
\begin{tikzpicture}[scale=2]
\clip (-0.1,-0.1) rectangle (1.3, 0.3);
\draw [gray, thick, line width = 1pt] (0.1,0.2) -- (0.1,0.25);
\draw [gray, thick, line width = 1pt] (0.2,0.2) -- (0.2,0.25);
\draw [gray, thick, line width = 1pt] (0.3,0.2) -- (0.3,0.25);
\draw [gray, thick, line width = 1pt] (0.5,0.2) -- (0.5,0.25);
\draw [gray, thick, line width = 1pt] (0.6,0.2) -- (0.6,0.25);
\draw [gray, thick, line width = 1pt] (0.7,0.2) -- (0.7,0.25);
\draw [gray, thick, line width = 1pt] (0.9,0.2) -- (0.9,0.25);
\draw [gray, thick, line width = 1pt] (1.0,0.2) -- (1.0,0.25);
\draw [gray, thick, line width = 1pt] (1.1,0.2) -- (1.1,0.25);

\node[] (a) at (0,0.2) {};
\node[] (b) at (0.4,0.2) {};
\node[] (d) at (1.2,0.2) {};
\node[] (c) at (0.8,0.2) {};
\draw[red, very thick] (0,0.2) circle(0.03cm);
\draw[green, very thick] (0.4,0.2) circle(0.03cm);
\draw[blue, very thick] (1.2,0.2) circle(0.03cm) ;
\draw[black, very thick] (0.8,0.2) circle(0.03cm);
\path[-,very thick, gray, line width = 1.2pt] 
(a) edge  (b)
(b) edge (c) 
(c) edge (d);  
\draw[->, very thick, gray, line width = 1.2pt] (d) -- (1.3, 0.2);
\node at (a) [black, below = 1pt]{$0$};
\node at (c) [black, below =1pt]{$r_{13}^q$};
\node at (b) [black, below =1pt]{$r_{12}^q$};
 \node at (d) [black, below =1pt]{$r_{14}^q$};
\end{tikzpicture}
} \\ (c) relation vector space \\using $v_1$ as a pivotal node
\end{tabular}
\\
\begin{tabular}{c}
{
\begin{tikzpicture}[scale=3]
\clip (0,0.05) rectangle (0.801,0.8);
\node[] (a) at (0.2,0.2) {};
\node[] (b) at (0.6,0.2) {};
\node[] (d) at (0.4,0.5) {};
\draw[red, very thick] (0.2,0.2) circle(0.03cm);
\draw[green, very thick] (0.6,0.2) circle(0.03cm);
\draw[black, very thick] (0.4,0.5) circle(0.03cm);
\path[-,very thick, gray, line width = 1pt] 
(a) edge node[black, below =-1pt]{$r_{12}$} (b)
(b) edge node[black, right =-1pt]{$r_{23}$}(d)
(a) edge node[black, above = -7pt]{$r_{13}$}(d); 
\node[below = 1pt] (a) at (0.2,0.2) { $v_1$};
\node[below = 1pt] (b) at (0.6,0.2) {$v_2$};
\node[left = 1pt] (d) at (0.4,0.5) {$v_3$};
\end{tikzpicture}
} \\ (a) data graph $G$
\end{tabular}
&
$\Rightarrow$
\begin{tabular}{c}
{\footnotesize
{
$\begin{array}{c|ccc}
        & v_1        & v_2       & v_3      \\ \hline 
v_1 & 0            & r_{12}   & r_{13}  \\
v_2 & r_{21}   & 0            & r_{23}   \\
v_3 & r_{31}   & r_{32}   & 0          \\[3pt]
\end{array} $
}
} \\ (b) adjacency matrix
\end{tabular}

&
$\Rightarrow$
\begin{tabular}{c}
{
\begin{tikzpicture}[scale=2]
\clip (-0.1,0) rectangle (1.3, 0.3);
\draw [gray, thick, line width = 1pt] (0.1,0.2) -- (0.1,0.25);
\draw [gray, thick, line width = 1pt] (0.2,0.2) -- (0.2,0.25);
\draw [gray, thick, line width = 1pt] (0.3,0.2) -- (0.3,0.25);
\draw [gray, thick, line width = 1pt] (0.5,0.2) -- (0.5,0.25);
\draw [gray, thick, line width = 1pt] (0.6,0.2) -- (0.6,0.25);
\draw [gray, thick, line width = 1pt] (0.7,0.2) -- (0.7,0.25);
\draw [gray, thick, line width = 1pt] (0.9,0.2) -- (0.9,0.25);
\draw [gray, thick, line width = 1pt] (1.0,0.2) -- (1.0,0.25);
\draw [gray, thick, line width = 1pt] (1.1,0.2) -- (1.1,0.25);

\node[] (a) at (0,0.2) {};
\node[] (b) at (0.4,0.2) {};
\node[] (d) at (1.2,0.2) {};
\node[] (c) at (0.8,0.2) {};
\draw[red, very thick] (0,0.2) circle(0.03cm);
\draw[green, very thick] (0.4,0.2) circle(0.03cm);
\draw[black, very thick] (0.8,0.2) circle(0.03cm);
\path[-,very thick, gray, line width = 1.2pt] 
(a) edge  (b)
(b) edge (c) 
(c) edge (d);  
\draw[->, very thick, gray, line width = 1.2pt] (1,0.2) -- (1.3, 0.2);
\node at (a) [black, below = 1pt]{$0$};
\node at (c) [black, below =1pt]{$r_{13}$};
\node at (b) [black, below =1pt]{$r_{12}$};
\end{tikzpicture}
} \\ (c) relation vector space\\ using $v_1$ as a pivotal node
\end{tabular}
\end{tabular}
\caption{Relation vector space to simplify relation assignment. In this illustration, it shows two different graphs: $Q$ and $G$, the corresponding adjacency matrices and relation vector spaces for a single pivotal node $v_1$. 
}\label{relationV}
\end{SCfigure*}
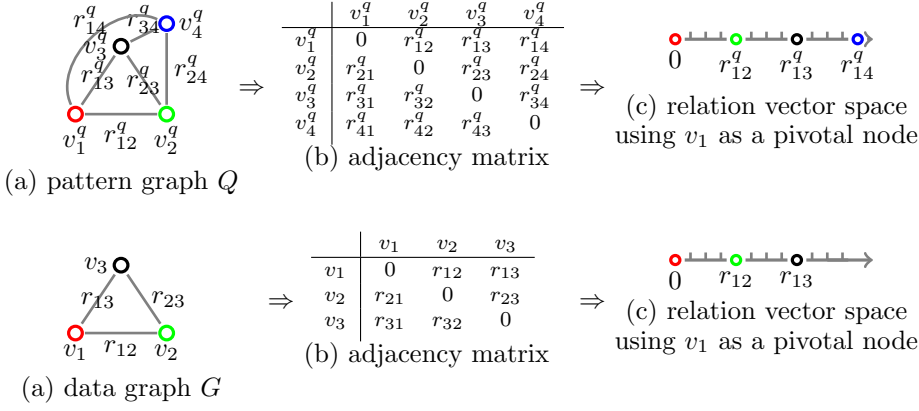

\subsection{Content extraction via graph mining} 
Given the pattern graph $Q$, to extract similar graphs from a document, it starts with pivotal nodes selection in a document and perform relation assignment to compute feature score between the pairs of nodes. Relations assignment repeats until a similar graph $G$ is achieved, with respect to $Q$.  

\bigskip
\noindent \textit{\textbf{Pivotal nodes selection.}} In a predefined set $ {\cal L}$ of labels such as \textit{price}, \textit{date}, \textit{address} and \textit{description} in the domain, for every node $v_i^q$ in pattern graph $Q$, the corresponding label $\ell_i^q\in \cal L$ is defined i.e., $V^q = \{ (v_i^q, \ell_i^q), i= 1 \ldots \mathbb V^q\}$. Having these labelled nodes $\big\{ (v_i^q, \ell_i^q) \big\}$ in a pattern graph $Q$, the target is to select nodes sharing identical labels $\big\{ (v_{\hat{i}}, \ell_{\hat{i}}) \big\}$ from a document $d$. We now, refer the selected nodes as pivotal nodes.

\bigskip
\noindent \textit{\textbf{Feature score computation.}}  Each pivotal node is taken and started to validate relations with neighbouring nodes in a document, as in pattern graph. To compute feature score between the pair of nodes $(v_i,v_j)$ in a document with respect to $(v_i^q, v_j^q) \in Q$, their respective relations  must be identical i.e., $r_{ij}^q$ validates with $r_{ij}$. More formally, we can compute feature score between two corresponding nodes $v^q$ and $v$ as $f.\mbox{score}(v^q,v) =$
\begin{eqnarray} \label{f.score}
\left\{
\begin{array}{l}
1: \mbox{ label in } v^q = \mbox{ label in } v\mbox{, and } \\
\frac{1}{\mathbb F} \sum_f \lambda_f \times  s_{v^q, v}^{\mbox{feature}_f}: \mbox{ otherwise,}
\end{array}\right.
\end{eqnarray}
where $\lambda_f \in [0,1]$ provides weight to each features used to compute feature matching score $s_{(,)}$. For each particular feature, weight $\lambda_f$ can be varied according to its robustness and so is application dependent. Given two strings: $x$ reference and $y$ primary, we compute feature (like string \textit{value}, number of \textit{word}s and \textit{size} (\textit{cf.} Eq.~\eqref{eq_field})) matching scores as follows. \\ 
\noindent $\bullet$ String \textit{type}: \\
$s_{x,y}^{type} = 1 - \big( {\mbox{Levenshtein dist.}(x,y)}/{\max(x,y)}\big)$, where we treat numerals $\{0-9\}$, all alphabets $\{A-Z, a-z\}$ and symbols equally. \\
\noindent $\bullet$ Number of \textit{word}s in a string: \\
$s_{x,y}^{word} = 1 - \big({\mbox{dist.}^{word}(x,y)}/{\max(x,y)}\big)$ i.e., an absolute difference in number words is normalised by the maximum number of words. \\
\noindent $\bullet$ String \textit{size}: \\
$s_{x,y}^{length} = 1 - \big({\mbox{dist.}^{length}(x,y)}/{\max(x,y)}\big)$ i.e.,  an absolute difference in size (number of letters) is normalised by its maximum size.\\

Following Fig.~\ref{relationV}, let us elaborate a concept of matching. To simplify the explanation, let us first create a relation vector space from a pattern graph and then realise the assignment process for each pivotal node in a document. Taking a single pivotal node $v_1$ from a data graph $G$ (having identical label with respect to $v_1^q$ in $Q$ i.e., $\ell_1^p = \ell_1^q \in \cal L$), the idea is to assign relations $\big\{r^q_{12}, r^q_{13}, r^q_{14}\big\}$ in data graph $G$. We validate relations $\{ r_{12}, r_{13} \}$ one-by-one and compute feature score in parallel. It provides $G \subseteq Q$. However, an addition of a node $v_3$ can help to make them exactly similar in configuration via an edit cost operation. 

\bigskip
\noindent \textbf{\textit{Graph matching score computation.}} 
An aggregation of both scores i.e., $r.$score from relation assignment and $f.$score from feature computation between the nodes yields a matching score $S$ for data graph $G$ with respect to $Q$ 
\begin{eqnarray}\label{matching score} 
S(Q,G) &=& \alpha \frac{1}{\mathbb R}\sum_{{i,j} \in {\cal R}^q, i\neq j} r.\mbox{score} (r^q_{i,j}, r_{i,j}) + \\ \nonumber
          && (1- \alpha) \frac{1}{\mathbb V^q} \sum_{i \in V^q} f.\mbox{score}(v^q_i, v_i), \alpha \in [0,1]. 
\end{eqnarray}

\bigskip
\noindent \textbf{\textit{Confidence score computation.}} From each input pattern, a set of mined graphs $\{ (G_g, S_g)\}$ will represent a table i.e., an output. For such an output, we compute corresponding confidence score (CS). CS is computed from the aggregation of all matching scores $\{ S_g\}_{g=1}^\mathbb G$, which is then normalised i.e., $\mbox{CS}_\Bbbk^{t_n} = \frac{1}{\mathbb{G}}\sum_{g=1}^{\mathbb{G}} S_g.$ 
In case of multiple input patterns, the outputs are ranked and provided on a one-to-one basis. Ranking is based on the order of similarity. 

Note that we aim to use set of mined graphs to iteratively update the pattern graph and transform into a graph model so that it can be used in the absence of the clients -- which is beyond the scope of the paper. A proof of the concept is reported in~\cite{kc:IBPRIA2013} and the thorough extension (aiming to apply document information content extraction, not necessarily be always found in structured documents like forms) has been made in~\cite{kc:ICDAR2013}.

\section{Experiments}\label{sec:expe}
\subsection{Dataset and evaluation metric} 
\noindent \textbf{\textit{Dataset.}} We work on a real-world industrial problem in direct collaboration with the \textbf{ITESOFT}\footnote{\url{http://www.itesoft.com}.}, France. Currently, the dataset is composed of 15 classes with 100 samples per class. 
For each document, clients provide ground-truths i.e., all similar patterns within the table, according to the pattern selected.

\bigskip
\noindent \textbf{\textit{Evaluation metric.}} An output i.e., the detected table is represented by a collection of mined graphs $O = \{ G_g, S_g\}$ in a test document, and there are $\mathbb G^\circ$ list of ground-truthed patterns corresponding to the ground-truthed table $O^\circ = \{ G^\circ_g\}_{g^\circ =1}^{\mathbb G^\circ}$. Each graph $G$ has a number of fields that are simply represented by iconic boxes $\{ B_b\}_{b = 1}^\mathbb B$. 

To evaluate, we extend the area-ratio-based measure proposed by Shafait and Smith~\cite{ShafaitS10:DAS}. It uses bounding boxes to describe detected tables and the ground-truths. In our framework, the overlapping ratio between the two boxes is defined as
$
OR_1(B_b^{\circ}, B_{b}) = \frac{2 \times  |B_b^{\circ} \cap B_{b}|}{ | B_b^{\circ} | +  | B_{b} | },
$
where $| B_b^{\circ} \cap B_b|$ is the intersected or common area of two bounding boxes from ground-truthed and detected table respectively and $| B_b^{\circ} |, |B_{b}|$ are the individual areas. Note that $OR_1(,) \in [0,1]$. We sum up all $OR_1(,)$ and normalise to compute overall overlapping ratio between ground-truth pattern  $G^\circ$ and detected pattern $G$ by $OR_2(G^{\circ},G) = \frac{1}{\max(\mathbb B^{\circ}, \mathbb B)} \sum OR_1(B_b^{\circ}, B_{b}), \{ b^{\circ}: b^{\circ}\in {\mathbb B^{\circ}} \wedge b \in {\mathbb B^{\circ}} \}$. Then for a whole table, we can express evaluation metric as
\begin{eqnarray}
Eval({O^{\circ}, O}) = & \frac{1}{\max(\mathbb G^{\circ}, \mathbb G)}\sum OR_2(G_g^{\circ}, G_{g}), \\ \nonumber
&\{ g^{\circ}: g^{\circ}\in {O^{\circ}} \wedge g \in {O^{\circ}} \}.
\end{eqnarray} 

\subsection{Results and analysis} 
We have validated the outputs over 15 different suppliers by taking the associated ground-truths and reported the average performance in Table~\ref{table: result}. More specifically, it provides the two different ways to evaluate: 
\begin{enumerate}\itemsep-0.5em
\item one is associated with the input pattern created in the laboratory and 
\item another one is directly related with client or real-world patterns. 
\end{enumerate}
The first evaluation of course, aims to provide an overall concept that can be applied to content extraction associated with the table. The latter one provides how robust it is. 
In the reported results in Table~\ref{table: results}, we observe the following. 
\begin{enumerate}\leftmargin = 0em\itemsep-0.5em 
\item Without a surprise, cleaner the input pattern, better the performance. This happens to be in \textit{eval. 1} since input patterns are created in accordance with what OCR results. 
\item In contrast, in case of the client input patterns (\textit{eval. 2}), a single field selection may sometimes take word(s) from another closer fields (can be left or right), and multiple lines. In that selected box (from clients), since OCR reads some dots (due to noise) as `full-stop', `colon' and `semi-colon', it does not allow possible cleaning. As a consequence, feature properties representing the graph nodes can possibly varied. Fig.~{\color{red}{6}}. shows an example of it. 
\end{enumerate}

\begin{table}[b]
\centering
\small
\caption{Average performance (in \%) over three different types of table: header, body and footer.}\label{table: results}\label{table: result}
\renewcommand{\tabcolsep}{0.7em}
\begin{tabular}{lcccc}
\hlinewd{1.5pt} 
Table type$_\Rightarrow$ & \rotatebox{0}{Header} & \rotatebox{0}{Body} & \rotatebox{0}{Footer} & \rotatebox{0}{Avg.} \\ \hlinewd{1pt} 
\textit{Eval. 1}		& 97	& 99  & 98 & 98 \\ 
\textit{Eval. 2}		& 96	& 98  & 95 & 97 \\ \hlinewd{1pt} 
\multicolumn{5}{l}{\textit{Eval. 1}: input patterns created in lab.}\\
\multicolumn{5}{l}{\textit{Eval. 2}: input patterns from clients.}\\
\multicolumn{5}{l}{Execution time $\simeq$ 2 sec./doc. image.} \\ \hlinewd{1.5pt} 
\end{tabular}
\end{table}
\begin{figure*}[t]
\centering
\small
\begin{tabular}{cc}
\begin{tikzpicture}
\node[-, draw, gray!50] (b) at (0,3.6) {\includegraphics[scale=0.5]{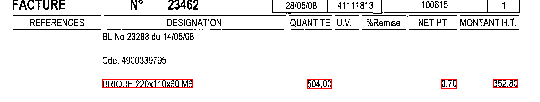}};
\node at (0.4, 3.6) {\color{red}{input pattern (linear)}};
\node[-, draw, gray!50](b) at (0.2,-2.0) {\includegraphics[scale=0.65]{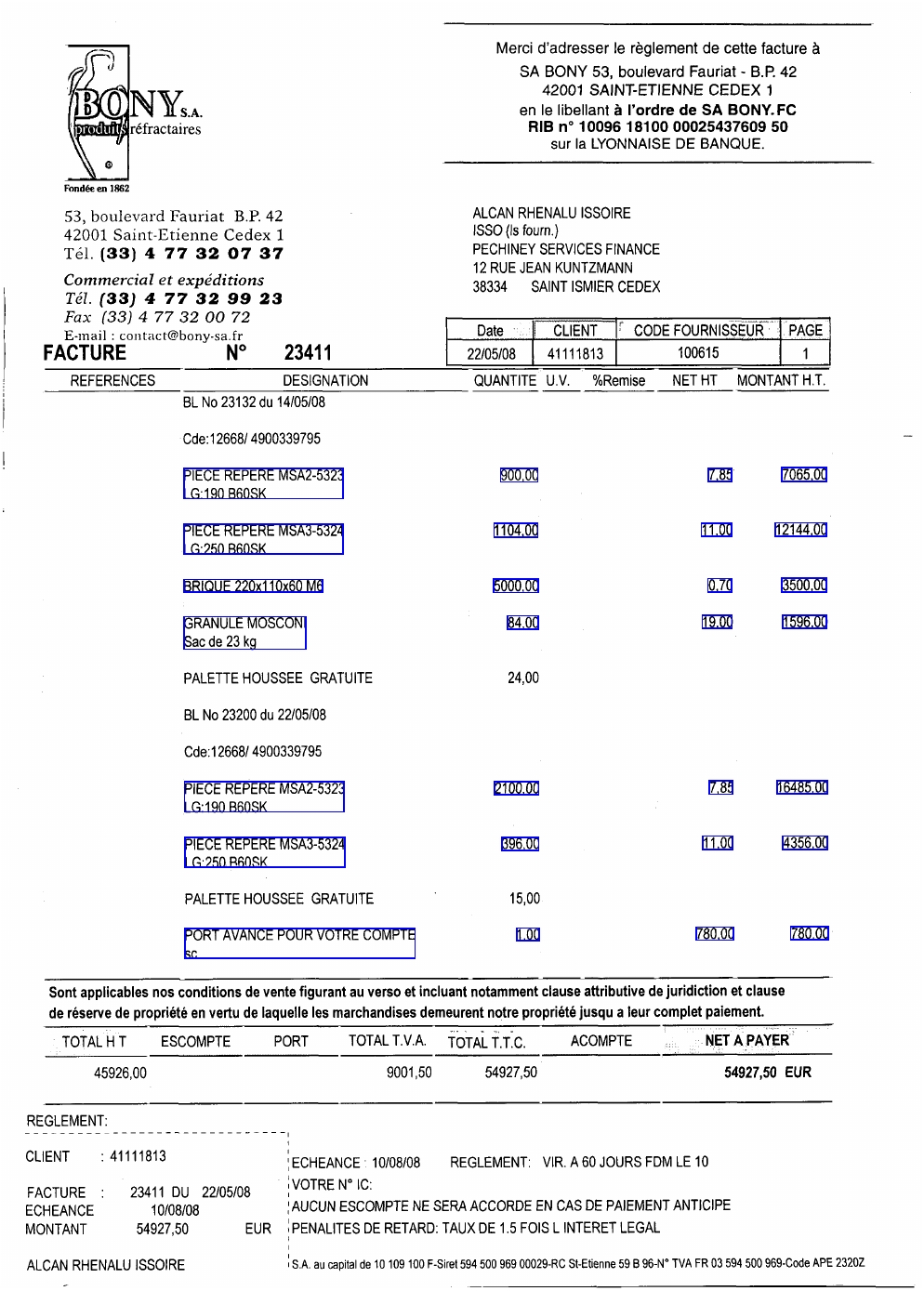}};
\node[-, draw, gray!50](b) at (0.2,-2.2) {\includegraphics[scale=0.65]{image/ENV490.pdf}};
\node[-, draw, gray!50](b) at (0.4,-2.4) {\includegraphics[scale=0.65]{image/ENV490.pdf}};
\node[-, draw, gray!50](b) at (0.6,-2.6) {\includegraphics[scale=0.65]{image/ENV490.pdf}};
\node at (2, -1.0) {\color{blue}{$\swarrow$output patterns}};
\node at (-3.5, -7.0) {(a)};
\node at (0.5, 1.0) {\huge \color{gray}{\textbf{CONFIDENTIAL}}};
\node at (-3.3, -3.0) {\rotatebox{90}{\color{black}{\textbf{list of doc. images + outputs}}}$_{\Longrightarrow}$};
\node at (-1.8, -1.48) {\color{magenta}{\textbf{1)}}};
\node at (-1.8, -1.88) {\color{magenta}{\textbf{2)}}};
\node at (-1.8, -2.25) {\color{magenta}{\textbf{3)}}};
\node at (-1.8, -2.6) {\color{magenta}{\textbf{4)}}};
\node at (-1.8, -3.75) {\color{magenta}{\textbf{5)}}};
\node at (-1.8, -4.15) {\color{magenta}{\textbf{6)}}};
\node at (-1.8, -4.75) {\color{magenta}{\textbf{7)}}};
\end{tikzpicture} 
&
\begin{tikzpicture}
\node[-, draw, gray!50] (b) at (0,3.6) {\includegraphics[scale=0.305]{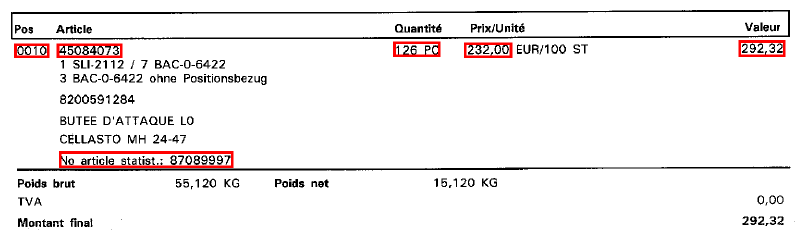}};
\node at (0.7, 3.4) {\color{red}{input pattern (zig-zag)}};
\node[-, draw, gray!50](b) at (0.2,-2.0) {\includegraphics[scale=0.11]{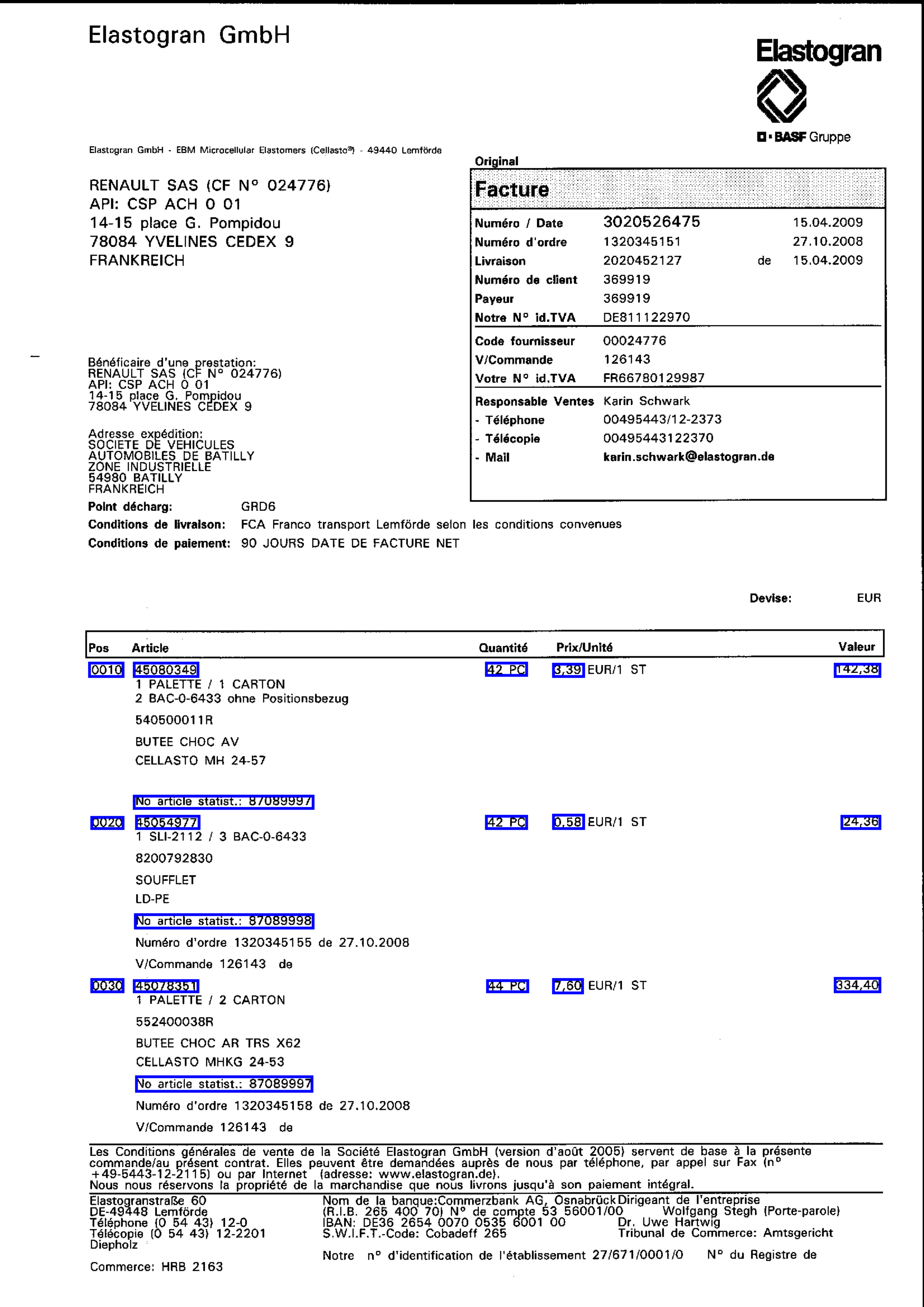}};
\node[-, draw, gray!50](b) at (0.2,-2.2) {\includegraphics[scale=0.11]{image/ENV65_1_1out.pdf}};
\node[-, draw, gray!50](b) at (0.4,-2.4) {\includegraphics[scale=0.11]{image/ENV65_1_1out.pdf}};
\node[-, draw, gray!50](b) at (0.6,-2.6) {\includegraphics[scale=0.11]{image/ENV65_1_1out.pdf}};
\node at (2, -2.0) {\color{blue}{$\swarrow$output patterns}};
\node at (-3.5, -7.0) {(b)};
\node at (0.5, 1.5) {\huge \color{gray}{\textbf{CONFIDENTIAL}}};
\node at (-3.3, -3.0) {\rotatebox{90}{\color{black}{\textbf{list of doc. images + outputs}}}$_{\Longrightarrow}$};
\node at (-2.5, -3.0) {\color{magenta}{\textbf{1)}}};
\node at (-2.2, -3.2) {\color{magenta}{$\Bigg\{$}};


\node at (-2.5, -4.0) {\color{magenta}{\textbf{2)}}};
\node at (-2.2, -4.2) {\color{magenta}{$\Bigg\{$}};


\node at (-2.5, -5.2) {\color{magenta}{\textbf{3)}}};
\node at (-2.2, -5.3) {\color{magenta}{$\Bigg\{$}};
\end{tikzpicture}
\end{tabular}
\mbox{}\\ \mbox{}\\
\caption{ Examples showing content extraction within the table in accordance with the input pattern (from client). Tables are composed of separately (a) seven and (b) three similar patterns in two different suppliers.}\label{example_tableExt}
\end{figure*}
\balance
Besides, another considerable issue is the complexity of the graph-based pattern representation. In case of input patterns with complex structural formats (lets say zig-zag), such non-selected fields integration makes pattern graph more complex. Furthermore, as said before, our system performance has been affected due to OCR errors since the system does not provide the expected semantics label at nodes in the graph. An example of the OCR effect is `false detection' because of the structural similarity between the graphs. 

\section{Conclusions and future perspectives}\label{sec:conc}
In this paper, we have presented client-driven pattern-based approach to table extraction via graph mining scheme, inspiring from a real-world applications. We have very much focused and validated that the table extraction does not always mean only to detect the presence and absence as well as to spot the area where table(s) is(are) located but also to select important key fields within it while rejecting others. 

Given an input pattern (i.e., a pattern graph), finding similar pattern graphs so that we can reinforce or update it iteratively each time we extract them, is one of the primary issues of the further work~\cite{kc:IBPRIA2013,kc:ICDAR2013}, for instance. As a consequence, such models are used to exploit document information content in the absence of clients.



\end{document}